\documentclass[lettersize,journal]{IEEEtran}
\usepackage{amsmath,amsfonts}
\usepackage{algorithmic}
\usepackage{algorithm}
\usepackage{array}
\usepackage[caption=false,font=normalsize,labelfont=sf,textfont=sf]{subfig}
\usepackage{textcomp}
\usepackage{stfloats}
\usepackage{url}
\usepackage{verbatim}
\usepackage{graphicx}
\usepackage{ulem}
\makeatletter
\let\em\relax
\DeclareRobustCommand\em{%
  \@nomath\em
  \ifdim \fontdimen\@ne\font >\z@
    \upshape
  \else
    \itshape
  \fi}
\let\emph\em
\makeatother

\usepackage{cite}
\usepackage{hyperref}
\usepackage{multirow}
\usepackage{graphicx} 
\usepackage{textcomp}
\usepackage{soul, color, xcolor}
\usepackage{orcidlink}
\usepackage{fancyhdr}
\begin{document}

\title{Multimodal Industrial Anomaly Detection via Geometric Prior}

\author{
Min Li$^{1,2,3\orcidlink{0000-0002-0507-5576}}$ , 
Jinghui He$^{2,3\orcidlink{0009-0003-2249-1250}}$ , 
Gang Li$^{2,3,1,*\orcidlink{0000-0002-7896-4833}}$ , 
Jiachen Li$^{2,3\orcidlink{0000-0002-3543-6088}}$ , 
Jin Wan$^{2,3\orcidlink{0000-0001-9245-0110}}$ 
and Delong Han$^{2,3\orcidlink{0000-0001-7195-3413}}$    
\thanks{$^{1}$ Faculty of Data Science, City University of Macau, Macau, China.  }%
\thanks{$^{2}$ Key Laboratory of Computing Power Network and Information Security, Ministry of Education, Shandong Computer Science Center (National Supercomputer Center in Jinan), Qilu University of Technology (Shandong Academy of Sciences), Jinan, China.}%
\thanks{$^{3}$ Shandong Provincial Key Laboratory of Computing Power Internet and Service Computing, Shandong Fundamental Research Center for Computer Science, Jinan, China.  }%
\thanks{This work is supported by the Key R\&D Program of Shandong Province, China (2024CXGC010108), the National Natural Science Foundation of China (62401305), the Taishan Scholars Program (NO.tsqn202103097, NO.tsqnz20240834), and the Qilu Youth Innovation Team (2024KJH028).}%
}


\IEEEpubid{\begin{minipage}{\textwidth}\ \\[30pt] \centering
		Copyright \copyright 2025 IEEE. Personal use of this material is permitted. 
		However, permission to use this material for any other purposes must \\ be obtained 
		from the IEEE by sending an email to pubs-permissions@ieee.org.
\end{minipage}}

\maketitle

\begin{abstract}
The purpose of multimodal industrial anomaly detection is to detect complex geometric shape defects such as subtle surface deformations and irregular contours that are difficult to detect in 2D-based methods. However, current multimodal industrial anomaly detection lacks the effective use of crucial geometric information like surface normal vectors and 3D shape topology, resulting in low detection accuracy. In this paper, we propose a novel Geometric Prior-based Anomaly Detection network (GPAD). Firstly, we propose a point cloud expert model to perform fine-grained geometric feature extraction, employing differential normal vector computation to enhance the geometric details of the extracted features and generate geometric prior. Secondly, we propose a two-stage fusion strategy to efficiently leverage the complementarity of multimodal data as well as the geometric prior inherent in 3D points. We further propose attention fusion and anomaly regions segmentation based on geometric prior, which enhance the model's ability to perceive geometric defects. Extensive experiments show that our multimodal industrial anomaly detection model outperforms the State-of-the-art (SOTA) methods in detection accuracy on both MVTec-3D AD and Eyecandies datasets.
\end{abstract}

\begin{IEEEkeywords}
Industrial Anomaly Detection, Multimodal Anomaly Detection, Geometry-Conditioned Attention.
\end{IEEEkeywords}

\section{Introduction}
\begin{figure}
    \centering
    \includegraphics[width=1\linewidth]{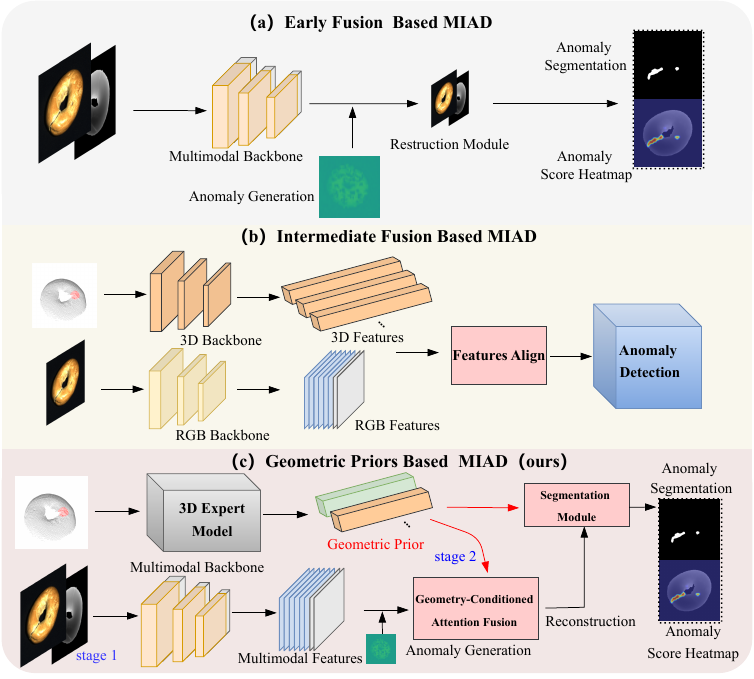}
    \caption{Comparison of three MIAD pipelines: (a) early fusion based, (b) intermediate fusion based, and (c) geometric prior based. Our framework integrates a dual fusion strategy, using a 3D expert model to extract geometric prior, which are then used to guide multimodal fusion and abnormal region segmentation.}
    \label{figure1}
\end{figure}
\IEEEPARstart{M}{ultimodal} industrial anomaly detection (\textbf{MIAD}) \cite{asad20252m3df,chu2023shape,wang2023multimodal,zavrtanik2024cheating,zhou2024pointad,cao2024complementary,wu2024cross} integrates data from multiple modalities, such as RGB images, depth images, and point clouds, to improve anomaly detection in industrial environments. Unlike traditional RGB-based anomaly detection methods \cite{li2023rethinking,zavrtanik2022dsr,huang2024attention,you2022adtr,roth2022towards,jiang2024fr,hu2024dmad,jiang2022softpatch,xing2023visual,yao2023learning,chen2024progressive} that depend on a single modality for defect detection. Real-world industrial scenarios require capturing abnormalities across various product attributes such as shape, surface texture, and internal structure, and multimodal data can provide a more comprehensive and accurate representation for detection \cite{wang2024real}. By using RGB images, depth images, and point clouds, MIAD has the potential to further improve the reliability and accuracy of anomaly detection in industrial environments \cite{liu2024deep,lin2024survey}.

However, existing MIAD methods \cite{asad20252m3df,chu2023shape,wang2023multimodal,zavrtanik2024cheating}  still exhibit several limitations. As illustrated in Fig.\hyperref[figure1]{1} (a), early fusion based MIAD methods \cite{farahnakian2022rgb} suffer from RGB-induced modality bias, where the high-dimensional color information of RGB tends to dominate over the depth information. Specifically, during backpropagation, the high-dimensional color information of RGB has a stronger influence on the parameter optimization direction through competitive gradient flows. In contrast, the low-bit depth maps are systematically suppressed in the subsequent convolution operations. This asymmetric modality encoding not only damages the geometric integrity of 3D structures due to indiscriminate feature aggregation but also leads to a self-reinforcing bias where the network gradually neglects depth features in deeper layers. Fig.\hyperref[figure1]{1} (b) demonstrates that intermediate fusion approaches \cite{wang2023multimodal,zhou2024pointad,rotstein2022multimodal} often encounter geometric feature distortion such as inconsistent shape representation caused by inadequate cross-modal alignment mechanisms. Specifically, the lack of explicit geometric constraints during feature interaction results in inconsistent representation spaces between RGB and point cloud modalities. This structural discrepancy not only causes geometric information loss during projection operations but also leads to asymmetric feature propagation paths, thus affecting the synergistic representation. Moreover, the modality-specific feature extractors with different architectures produce features with divergent statistical distributions, which worsen the misalignment problems and reduce the detection sensitivity for geometric anomalies.

To address these limitations, our GPAD framework introduces geometric prior-guided hierarchical fusion as shown in Fig.\hyperref[figure1]{1} (c), which aims to improve the alignment and integration of multimodal features. The first stage establishes a geometry-aware RGB-D representation through channel-wise concatenation of raw depth and RGB data. The explicit injection of the depth map provides a coarse geometric basis to prevent RGB from dominating. Based on this enhanced multimodal foundation, the second stage uses a dedicated point cloud expert model to extract fine-grained geometric prior. Surface normal vectors are calculated through differential coordinate analysis to capture shape variations at the micrometer level. These geometric prior then drive an adaptive projection module that dynamically aligns 3D structural features with the RGB-D feature space, effectively resolving the cross-modal misalignment observed in conventional intermediate fusion methods. In particular, our fusion mechanism employs geometry-conditioned attention that utilizes normal vector as positional constraints during feature interaction. This enables geometry-aware feature recombination, where texture and geometric features are fused in a proportion that is related to their local shape complexity, thus enhancing the model's ability to perceive geometric imperfections.

Our contributions are summarized as follows:
\begin{itemize}
    \item We propose GPAD, a novel multimodal industrial anomaly detection method, which improves detection ability of structural anomalies by combining geometric prior.
    \item We propose a pre-trained point cloud expert model to perform fine-grained geometric feature extraction, using normal vectors to accurately capture the geometric details of features and generate geometric prior for guiding the subsequent fusion and segmentation processes.
    \item We propose a two-stage fusion strategy that not only effectively exploits the complementarity of multimodal data but also enhances the geometric perception ability of the model through the use of geometric prior in the fusion process.
    \item We propose to use geometric prior to guide geometry-conditioned attention fusion and abnormal region segmentation, and achieve higher detection and segmentation accuracy than the state-of-the-art methods on the MVTec-3D AD dataset.
\end{itemize}

\section{Related Work}
\label{sec2}

To address the diverse anomaly types in complex industrial scenarios, researchers have developed various detection methods based on RGB images, depth images, and point clouds, achieving significant advances in both single-modality and multimodal fusion approaches.

\subsection{RGB-based Industrial Anomaly Detection}

The mainstream RGB-based industrial anomaly detection methods can be classified into three main categories: reconstruction-based, memory bank-based, and diffusion model-based approaches \cite{li2023rethinking,zavrtanik2022dsr,you2022adtr,roth2022towards,liu2024deep,jiang2024fr,jiang2022softpatch,dai2024generating,zhang2023diffusionad,fuvcka2025transfusion}.

Reconstruction-based methods \cite{liu2024deep,lin2024survey} detect anomalies by learning from normal samples and reconstructing inputs during detection, where reconstruction errors serve as indicators of potential anomalies. MOOD \cite{li2023rethinking} enhances the model's understanding of the normal data distribution by reconstructing the masked region of the image, thereby improving the detection of abnormal regions. DSR \cite{zavrtanik2022dsr} uses a feature-quantized bisubspace reprojection network and a dual decoder mechanism to achieve a better distinction between normal and abnormal samples in feature space. AGPNet \cite{huang2024attention} uses an attention-guided perturbation mechanism to enhance the reconstruction of local features of an image through noise to improve the sensitivity and accuracy of detection. ADTR \cite{you2022adtr} is based on the reconstruction of Transformer features and achieves anomaly detection by learning differential features, effectively avoiding the problem of \textquotedblleft isomorphic mapping \textquotedblright. Overall, reconstruction-based image anomaly methods require self-trained encoders and decoders, which reduces the reliance on pre-trained models. This reduces reliance on pre-trained models but makes it difficult to extract advanced semantic features \cite{liu2024deep}.

Memory bank-based methods \cite{liu2024deep,lin2024survey} detect anomalies by comparing input features with a library of features from normal samples, leveraging the stored information for anomaly detection. Approaches like PatchCore \cite{roth2022towards}, FR-PatchCore \cite{jiang2024fr}, DMAD \cite{hu2024dmad}, and SoftPatch \cite{jiang2022softpatch} enhance detection with alignment, dual memory structures, and patch-level filtering. Although these methods are effective for rapid model development, they rely on pre-trained networks and require extra storage space for the feature libraries.

Diffusion model-based methods \cite{liu2024deep,lin2024survey} detect anomalies by learning the distribution of normal data through progressive generation or denoising, identifying deviations from the learned distribution. Examples include GRAD \cite{dai2024generating}, DiffusionAD \cite{zhang2023diffusionad}, TransFusion \cite{fuvcka2025transfusion} and AnomalySD \cite{yan2024anomalysd}. However, these methods typically require a large amount of computational resources, which limits their applicability in real-time scenarios \cite{livernoche2023diffusion}.

Although these methods \cite{li2023rethinking,zavrtanik2022dsr,you2022adtr,roth2022towards,liu2024deep,jiang2024fr,jiang2022softpatch,dai2024generating,zhang2023diffusionad,fuvcka2025transfusion} have made significant progress, challenges still exist in detecting complex industrial defects, particularly those that are partially hidden. Nevertheless, these methods have established a solid foundation for visual anomaly detection in industrial environments.

\subsection{3D-based Industrial Anomaly Detection}

3D anomaly detection methodologies \cite{liu2024deep,lin2024survey} leverage geometric-rich 3D representations (point clouds, depth maps, voxels) to capture spatial structures, offering inherent advantages over 2D approaches in industrial metrology applications.

3D-ST \cite{bergmann2023anomaly} employs a teacher-student architecture where the teacher network generates local geometric descriptors through iterative farthest point sampling, while the student network learns to predict descriptor distributions. IMRNet \cite{li2024towards} proposes a 3D anomaly synthesis pipeline to create a synthetic dataset called Anomaly-ShapeNet and introduces a self-supervised iterative mask reconstruction network for scalable representation learning and anomaly localization in 3D point clouds. R3D-AD \cite{liu2024real3d} introduces a large-scale high-precision point cloud anomaly detection dataset with high-resolution 3D items, and proposes a registration-based detection method with a novel feature memory bank.

While 3D methods \cite{bergmann2023anomaly,li2024towards,liu2024real3d} demonstrate superior geometric modeling capabilities, they exhibit critical limitations in industrial defect detection scenarios. The inability of geometric descriptors to capture surface reflectance characteristics and texture patterns makes some defects difficult to detect in industrial scenarios  \cite{tu2025self}. This limits the application of 3D-based methods in actual industrial quality control.

\subsection{Multimodal Industrial Anomaly Detection}

Multimodal approaches have emerged to overcome the limitations of single-modality methods by integrating information from multiple sources such as RGB images, depth images, and point clouds. This integration aims to improve the accuracy and robustness of anomaly detection in complex industrial environments. Existing MIAD methods can be categorized into early fusion, intermediate fusion, and late fusion based on the fusion stage \cite{baltruvsaitis2018multimodal,ramachandram2017deep}.

Early fusion based methods process multimodal data through sensor-level concatenation at the input stage. 3DSR \cite{zavrtanik2024cheating} introduces a simulation process to address the lack of diverse industrial depth datasets and uses a depth-aware discrete autoencoder (DADA) for 3D surface anomaly detection. But the concatenation of input layers in early fusion causes the depth information to be weakened in the deep network. Intermediate fusion based methods implement cross-modal interaction at intermediate feature levels. CFM \cite{costanzino2024multimodal} presents a novel cross-modal feature mapping framework which leverages point clouds and RGB images to localize anomalies by learning feature mappings between modalities on nominal samples. 2M3DF \cite{asad20252m3df} integrates multi-view RGB images and point clouds to capture global semantic context, local geometric structure and color information. Although intermediate fusion retains modality-specific features, the geometric distortion generated during feature projection affects the detection sensitivity. Late fusion based methods combine modality-specific predictions at the decision level. M3DM \cite{wang2023multimodal} fuses anomaly scores from separate memory banks through learnable weights. Although late fusion is computationally efficient, it lacks potential cross-modal synergy and has difficulty handling anomalies in the combination of geometry and texture.

Multimodal data fusion can combine the advantages of different modal data and make up for the shortcomings of a single modality. Multimodal anomaly detection methods in other fields, such as EGO fusion \cite{dinglearnable} and LAD-Reasoner \cite{li2025lad}, also aim to enhance anomaly detection performance through multimodal data fusion. EGO fusion \cite{dinglearnable} stands out by extending graph operations to merge multimodal features. It employs relationship graphs to capture feature interactions across different levels and leverages learnable graph fusion operators to dynamically integrate these relationships. On the other hand, LAD-Reasoner \cite{li2025lad} emphasizes the design of lightweight models, focusing on improving the efficiency and effectiveness of logical reasoning within the model. In contrast, our GPAD method innovatively uses geometric prior to guide the fusion of multimodal features. It highlights the extraction of fine-grained geometric features through specialized point cloud expert models and further utilizes these geometric prior to direct attention mechanisms, thereby establishing meaningful connections between features.

Current MIAD methods primarily use RGB modality as the foundation for identifying surface deformations and contour irregularities. However, existing approaches \cite{asad20252m3df,zavrtanik2024cheating,wang2023multimodal,zhou2024pointad,chu2023shape} face challenges such as modal competition, feature projection distortion, lack of cross-modal collaboration, and limited perception of geometric defects. GPAD introduces a two-stage fusion mechanism to address these issues. This mechanism effectively prevents modal competition and achieves cross-modal fusion using geometric prior, reducing geometric information loss and strengthening multimodal collaboration. In GPAD, we combine the geometric prior from 3D point clouds with the rich texture information from images. During feature interaction, the normal vector direction serves as a positional constraint, ensuring texture and geometric features are fused in proportion to their local shape complexity. This enhances the model's ability to perceive geometric defects, enabling more accurate detection of anomalies.


\begin{figure*}
    \centering
    \includegraphics[width=1\linewidth]{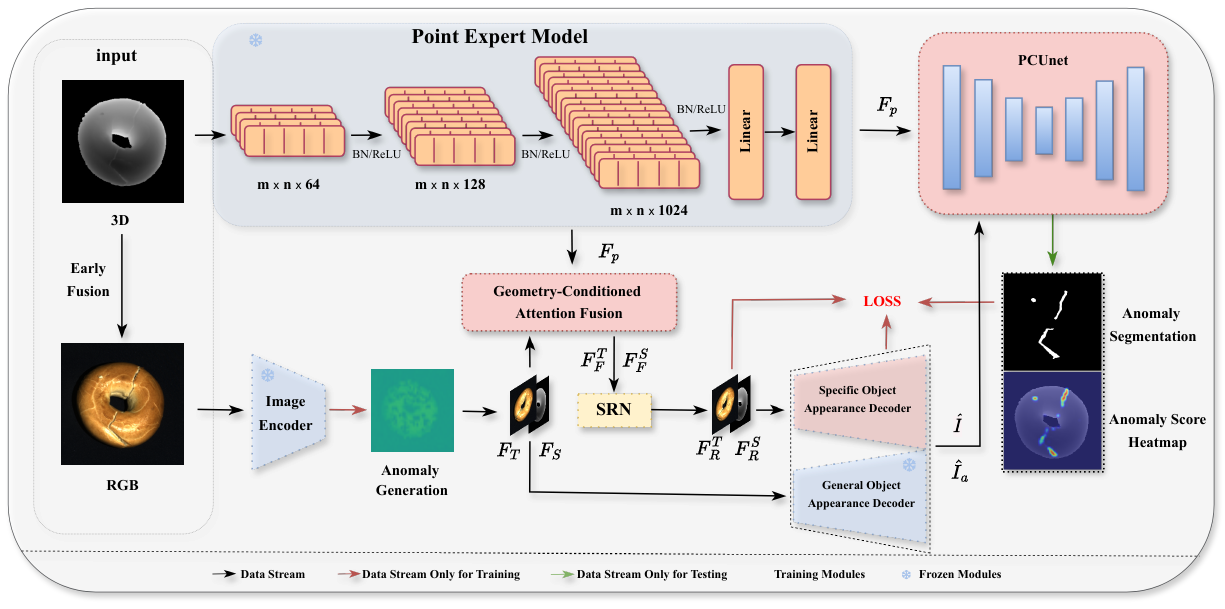}
    \caption{Architecture of GPAD. RGB-D features are extracted and quantized through a pretrained encoder, capturing both color and texture information as features $\mathbf\mathit{F_T}$, as well as high-level semantic features $\mathbf\mathit{F_S}$. Point clouds and their normals are processed by a point cloud expert model to extract fine-grained geometric features $\mathbf\mathit{F_p}$. These features are then fused with RGB-D features $\mathbf\mathit{F_T}$ and $\mathbf\mathit{F_S}$ via geometry-conditioned attention, fully leveraging the complementary information between modalities. The SRN module refines multimodal features, removing redundant information and enhancing anomaly detection through image reconstruction. $\hat{I}$ and $\hat{I}_a$ are the reconstructed images without and with anomalies, respectively. PCUnet employs geometric prior to guide the precise segmentation of anomaly regions.}
    \label{figure2}
\end{figure*}

\section{Method}

We introduce a novel MIAD framework, GPAD, a hierarchical fusion framework that establishes geometric consistency across modalities through differential normal analysis. GPAD consists of three main modules: a multimodal feature extraction module (Section \hyperref[3.2]{C}), a geometric prior-guided cross-modal fusion module (Section \hyperref[3.3]{D}), and an anomaly region segmentation module (Section \hyperref[3.4]{E}), where the loss function of the model training process is further detailed in Section \hyperref[3.5]{F}.

\subsection{Overview}

As shown in Fig.\hyperref[figure2]{2}, the GPAD framework processes RGB-D inputs and generates anomaly detection and segmentation results. 
Firstly, for the early-fused RGB-D image, GPAD employs an ImageNet-pretrained image encoder and a general-object appearance decoder \cite{liu2024deep,zavrtanik2022dsr,zavrtanik2024cheating}. This is done to address the common limitation of missing high-level semantic features \cite{liu2024deep} in reconstruction-based anomaly detection methods. 

During the training phase, anomalies are synthesized by selectively replacing feature embeddings with randomly sampled memory prototypes \cite{gong2019memorizing}, guided by a strength parameter and spatial constraints, while preserving the original structure through a mask-based blending mechanism. GPAD uses the pre-trained Depth-Aware Discrete Autoencoder (DADA) \cite{zavrtanik2024cheating} to extract low-level and high-level features of RGB-D images. After quantizing the features, it uses random replacement of feature embedding vectors to generate simulated abnormal features. Secondly, by converting the depth image into a point cloud and normal vector representation, GPAD extracts geometric prior through a 3D point cloud expert model and introduces a geometry-constrained cross-modal attention mechanism for fusion with RGB-D features. At the same time, it retains high-frequency geometric details through attention weight modulation. The RGB-D features and geometric prior are adaptively fused within the feature space via the geometry-conditioned attention fusion module, effectively integrating the unique characteristics of multimodal data. To further optimize the feature space representation \cite{liu2018linear}, we incorporate the subspace restriction network (SRN) module \cite{liu2018linear,zavrtanik2022dsr,zavrtanik2024cheating}, which constrains feature subspaces to reduce redundant information and noise, enhancing the relevance and expressiveness of multimodal features for anomaly detection and segmentation tasks. Finally, GPAD employs a dual-decoder structure \cite{zavrtanik2022dsr,zavrtanik2024cheating} for reconstruction, capturing fine details of multimodal features, and utilizes PCUnet for precise segmentation of anomaly regions. Specifically, the segmentation network enhances the geometric perception of abnormal areas by injecting geometric prior at multiple levels in the decoding stage.

\subsection{Notation Convention}
To facilitate understanding, we define the main symbols used in the method section as follows. We use $I_R$ and  $I_D$ to denote the RGB image and depth image, respectively. The point cloud is represented by $P_c \in \mathbb{R}^{M \times N \times 6}$, where M ($\text{M}$ = 200) represents the number of groups and N ($\text{N}$ = 500) represents the number of points contained in each group. Each point in the point cloud is obtained by concatenating its corresponding coordinates $p_i(\mathrm{x_i}, \mathrm{y_i}, \mathrm{z_i})$ and the corresponding normal vector $\mathbf{n_i}(\mathrm{n_{ix}}, \mathrm{n_{iy}}, \mathrm{n_{iz}})$. The geometric prior extracted by the Point Cloud Expert model is denoted by $\mathbf\mathit{{F}_p}\in\mathbb{R}^{M\times 128}$. The texture features and semantic features extracted from the pre-trained encoder of RGB-D data are denoted as $\mathbf\mathit{F_T}\in\mathbb{R}^{96\times 96\times d}$ and $\mathbf\mathit{F_S}\in\mathbb{R}^{48\times 48\times d}$ after introducing defects, which are used to reconstruct the anomaly image $\hat{I_a}$. After fusing $\mathbf\mathit{F_T}$ and $\mathbf\mathit{F_S}$ with the weighted geometric prior $F_W^T \in \mathbb{R}^{(48 \times 48) \times M}$ and $F_W^S \in \mathbb{R}^{(48 \times 48) \times M}$ respectively, the resulting features are denoted by $\mathbf\mathit{F_{F}^{T}}$, $\mathbf\mathit{F_{F}^{S}}\in\mathbb{R}^{48\times 48\times d}$. Finally, we pass ${F_{F}^{T}}$ and ${F_{F}^{S}}$ through the Subspace Restriction Network (SRN) to obtain $F_R^{T} \in \mathbb{R}^{ 96\times 96 \times d}$ and $F_R^{S}\in \mathbb{R}^{48\times 48 \times d}$, which are used to reconstruct the anomaly-free image $\hat{I}$.

\subsection{RGB-D Feature and Geometric Prior Extraction}
\label{3.2}

Multimodal feature extraction is a core component of GPAD, responsible for extracting essential features from RGB-D and point cloud data. This process provides comprehensive information for the subsequent geometry-guided attention fusion and segmentation modules.

\subsubsection{RGB-D Feature Extraction}
In the multimodal feature extraction process, RGB images supply the model with rich color and texture information, while depth images add spatial depth, capturing additional geometric details of the objects. In order to effectively utilize both types of information, RGB images $I_R$ and depth images $I_D$ are concatenated along the channel dimension. The DADA\cite{vu2019dada,zavrtanik2024cheating} module is then applied to extract features. Firstly, some initial convolutions are used to capture both color-texture features. Secondly, high-level semantic features are extracted through further convolutions and downsampling operations. Then, we perform feature quantization \cite{wu2016quantized,zavrtanik2022dsr,zavrtanik2024cheating} to reduce redundant information and lower computation and storage requirements. We introduce simulated abnormal features into both the quantized low-level features and high-level features to obtain $F_T \in\mathbb{R}^{96\times 96\times d}$ and $F_S \in\mathbb{R}^{48\times 48\times d}$, respectively.

\subsubsection{Geometric Prior Extraction}
Although RGB-D features $F_T$ and $F_S$ provide color, texture, and some geometric information, their 2D nature limits the ability to fully capture the 3D structure of objects. Point cloud data, which consists of sparse 3D points, conveys the object’s overall geometric shape and structural details, enabling the model to more accurately capture 3D structure.

Since the point cloud is large in size, in order to reduce the computational overhead, we first use Farthest Point Sampling (FPS) \cite{qi2017pointnet++,wang2023multimodal,chu2023shape} to divide the point cloud into M groups, each group contains N points, and the point cloud after division is denoted as $P \in \mathbb{R}^{M \times N \times 3}$. Each point is represented as $p_i\left( \mathrm{x_i}, \mathrm{y_i}, \mathrm{z_i} \right)$. Due to the inherent sparsity of point cloud data, capturing surface details can be challenging. To address this issue, we incorporate point cloud normals to enrich geometric feature information. Surface normals are computed through covariance analysis of local neighborhoods $\mathcal{N}(p_i)$, as shown in Equation (1):
\begin{equation}
\mathbf{n_i} = \arg\min_{|v|=1} v^\top \left( \sum_{p_j\in\mathcal{N}(p_i)}(p_j-\mu)(p_j-\mu)^\top \right) v,
\end{equation}
where $v$ is the unit normal vector, $\mu$ is the centroid of $\mathcal{N}(p_i)$. Then, a normal vector \cite{qi2017pointnet} is estimated for each point $\mathbf{n_i}(\mathrm{n_{ix}}, \mathrm{n_{iy}}, \mathrm{n_{iz}})$. The position coordinates and normal vectors are encoded together, forming an extended feature vector $\mathbf{{p}_{in}}$ = $\left[ \mathrm{x_i},\mathrm{y_i},\mathrm{z_i},\mathrm{n_{ix}},\mathrm{n_{iy}},\mathrm{n_{iz}} \right]$ for each point. This includes both spatial positions and normal vectors, forming an enriched input representation that facilitates the learning of geometric features by the subsequent network.

Due to the sparse convolution-based method and the complex operations of graph-based networks when processing point clouds, as well as the requirements of real-time in the industrial field. Our Point Cloud Expert (PCExpert) model implements hierarchical geometric coding through 1D convolution operations, as described in Fig.\hyperref[figure2]{2}.

After concatenating the coordinates and normal vectors of the points, we obtain the input of the point cloud expert model $P_c = \{P_c^1,P_c^2,...,P_c^M\}$, $P_c^i \in \mathbb{R}^{ N \times 6}$ is the $i$-th point cloud group. The PCExpert model uses a series of 1D convolutional layers, batch normalization, and ReLU activation functions to progressively extract group features, thereby enhancing spatial expressiveness. A global max pooling layer aggregates these local features into a compact global representation. As shown in Fig.\hyperref[figure2]{2}, the process of geometric prior extraction can be described as:
\begin{equation}
F_p^i =  \text{PCExpert}(P_c^i),
\end{equation}
where $F_p^i \in\mathbb{R}^{128}$ denotes the geometric prior of the $i$-th group and $P_c^i \in \mathbb{R}^{ N \times 6}$ is the $i$-th group of point cloud. The point coordinates provide the spatial positions information of the object's surface, while its corresponding normal vector provides directional information at each point. The PCExpert model learns to extract geometric features from the combination of positions and normals, which enables sensitivity to local shape variations. Surface defects, which induce abrupt changes in local geometry, are captured by these learned features. By aggregating the point cloud features of each group, we obtain the geometric prior ${F}_p\in\mathbb{R}^{M\times 128}$ for each sample.

\subsection{Geometry-Conditioned Attention Fusion}
\label{3.3}
\begin{figure}
    \centering
    \label{figure3}
    \includegraphics[width=1\linewidth]{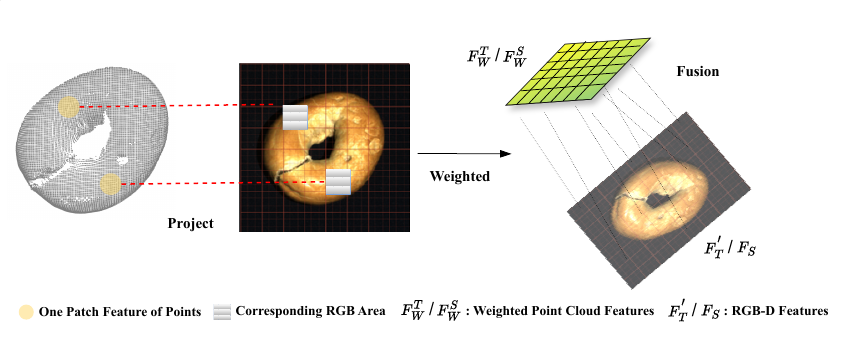}
    \caption{Geometry-Conditioned Cross-Modal Fusion. Each group of point cloud features is projected to the corresponding region in the image. The weighted point cloud features $F_W^T$ and $F_W^S$ are fused with $F^{'}_T$ and $F_S$, respectively.}
\end{figure}

The significant differences in distribution and spatial representation between RGB-D and point cloud features can result in information redundancy or loss when they are directly concatenated. This diminishes the complementary benefits of multimodal data. To address this, we design a geometry-conditioned attention fusion module to facilitate effective feature integration. As shown in Fig.\hyperref[figure3]{3}, this module takes RGB-D features as input, and fuses them with projected and weighted geometric prior, thus achieving efficient feature complementarity.

Firstly, we interpolate the texture features $F_T  \in\mathbb{R}^{96\times 96\times d}$ to the same dimension as the semantic features $F_S \in\mathbb{R}^{48\times 48\times d}$ by the following Equation:
\begin{equation}
\small
    F^{'}_T = \operatorname{Interpolate}\left(F_T, \text{size}=(48,48),\ \text{mode}=\text{bilinear}\right),
\end{equation}
where $F^{'}_T \in \mathbb{R}^{48 \times 48 \times d}$ represents the low-level texture features after interpolation, and bilinear indicates the use of bilinear interpolation.

At the same time, for each group's geometric prior $F_p^i \in\mathbb{R}^{128}$, we project it into the plane, an affine transformation is applied:
\begin{equation}
\hat{F}_p^i = \mathbf{\mathit{W_p}} F_p^i + b_p,
\end{equation}
where $\hat{F}_p^{i}\in \mathbb{R}^{(48 \times 48)}$ is the feature of each point cloud group after projection, $W_p \in \mathbb{R}^{(48 \times 48) \times 128}$ is the projection matrix shared across groups and $b_p \in \mathbb{R}^{(48 \times 48)}$ is the offset. We aggregate each point cloud group feature $\hat{F}_p^{i}$ to obtain the projected feature $\hat{F}_p \in \mathbb{R}^{ M \times (48 \times 48) }$ of each point cloud sample. This group-wise projection preserves local geometric structures and achieves dimensional consistency. By ensuring the projected point cloud features align with the spatial representation of RGB-D features, the projection enables a meaningful comparison and integration of features from different modalities in a shared feature space. 

To further exploit the geometric information from point clouds and the color-texture information from images, we introduce a geometry-conditioned attention mechanism, which dynamically generates attention weights for the geometric features.

Specifically, a linear transformation is applied to the projected point cloud features $\hat{F}_p$, and attention weights are calculated through a softmax function. Taking the fusion of texture features $F^{'}_T \in \mathbb{R}^{48 \times 48 \times d}$ and projected geometric prior $\hat{F}_p \in \mathbb{R}^{ M \times (48 \times 48) }$ as an example, the geometry-aware attention weights are computed as follows:
\begin{equation}
    \scriptsize
    \alpha_p^T = \text{Softmax}\left(\frac{(W_Q \times \text{Flatten}(F^{'}_T)^\top)^\top(W_K \times (\hat{F}_p)^\top)}{\sqrt{d}} + \Phi(\mathbf{n_i})\right),
\end{equation}
where $\alpha_p^T \in \mathbb{R}^{(48 \times 48) \times M}$ is the weight of the projected feature $\hat{F}_p$, $W_Q \in \mathbb{R}^{d \times d}$ and $W_K \in \mathbb{R}^{d \times (48 \times 48)}$ project features into a shared query-key space, $\Phi(\mathbf{n_i})$ represents the effect of normal vector encoding, and $d$ represents the feature dimension. Similarly, we can obtain the weight $\alpha_p^S \in \mathbb{R}^{(48 \times 48) \times M}$ of $\hat{F}_p$ to fuse with semantic features $F_S$.

Next, the $\hat{F}_p$ are modulated by the attention weights $\alpha_p^T$ and $\alpha_p^S$ through element-wise multiplication. The weighted geometric prior corresponding to $\alpha_p^T$ and $\alpha_p^S$ are obtained by the following Equations:
\begin{equation}
    F_W^T = (\alpha_p^T)^\top \odot \hat{F}_p,
\end{equation}
\begin{equation}
    F_W^S = (\alpha_p^S)^\top \odot \hat{F}_p,
\end{equation}
where $F_W^T \in \mathbb{R}^{(48 \times 48) \times M}$ represents the weighted geometric prior to be fused with $F^{'}_T \in \mathbb{R}^{48 \times 48 \times d}$, and $F_W^S \in \mathbb{R}^{(48 \times 48) \times M}$ represents the weighted geometric prior to be fused with $F_S \in \mathbb{R}^{48 \times 48 \times d}$. The attention mechanism dynamically adjusts the contribution of each point cloud group based on its geometric relevance to the corresponding image region. By leveraging the geometric prior to guide the attention mechanism, the model achieves a more precise and context-aware fusion of multimodal data, which is crucial to accurately identify and localize anomalies in industrial settings.

To yield the final fused representation, we first reshape the weighted geometric prior $F_W^T \in \mathbb{R}^{(48 \times 48) \times M}$ and $F_W^S\in \mathbb{R}^{(48 \times 48) \times M}$ into $\mathbb{R}^{48 \times 48 \times M}$, then use 1×1 convolution to align their channel dimension to $d$ respectively, and finally add them element-wise to the corresponding RGB-D features:
\begin{equation}
F_F^T = Conv_{1\times 1}(\text{Reshape}(F_W^T)) + F^{'}_T,
\end{equation}
\begin{equation}
F_F^S = Conv_{1\times 1}(\text{Reshape}(F_W^S)) + F_S,
\end{equation}
where $F_F^T \in\mathbb{R}^{48\times 48\times d}$ and $F_F^S \in\mathbb{R}^{48\times 48\times d}$ are the final multimodal robust feature representations. Guided by geometric prior, this attention fusion approach enables GPAD to effectively integrate geometric information from point clouds with color-texture features from images, significantly enhancing the accuracy and robustness of anomaly detection and segmentation in complex industrial scenarios.

\subsection{Anomaly Segmentation Guided by Geometric Prior}
\label{3.4}

GPAD employs a reconstruction-based anomaly detection method. For multimodal features $F_{F}^{T}$ and $F_{F}^{S}$, we then utilize the Subspace Restriction Network (SRN) \cite{liu2018linear,zavrtanik2022dsr,zavrtanik2024cheating} to perform multi-layer convolution and nonlinear transformation. This projects the multimodal features into a restricted feature space, capturing the basic structure of the data while preserving key geometric properties, and we obtain the corresponding feature representations $F_R^{T} \in \mathbb{R}^{ 96\times 96 \times d}$ and $F_R^{S}\in \mathbb{R}^{48\times 48 \times d}$. A dual-decoder strategy \cite{zavrtanik2022dsr, zavrtanik2024cheating} is implemented to accurately reconstruct both global structures and local details. The general decoder restores the overall visual features and ensures structural integrity through $F_T \in\mathbb{R}^{96\times 96\times d}$ and $F_S \in\mathbb{R}^{48\times 48\times d}$. Specifically, $F_S$ is upsampled to match the resolution of $F_T$, and the two are concatenated to serve as input to the general decoder. The reconstructed image $\hat{I}_a$ is as close as possible to the input image in appearance, but retains the anomalies generated in the image. Meanwhile, $F_R^{S}$ is upsampled and concatenated with $F_R^{T}$ as the input of the speific decoder to reconstruct $\hat{I}$. The anomalies in $\hat{I}$ will be removed as much as possible and reconstructed to a normal appearance.

\begin{figure}[t]
    \centering
    \includegraphics[width=1\linewidth]{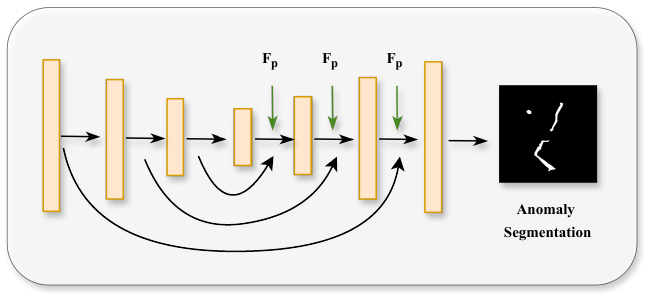}
    \caption{PCUnet architecture with geometric fusion. The geometric prior $F_p$ is adaptively projected and injected in each decoder layer to enhance the geometry awareness of the segmentation network.}
    \label{PCUnet}
\end{figure}

\begin{table*}
\label{table1}
    \centering
    \caption{Comparison of GPAD and Other Models on the MVTec3D Dataset (I-AUROC\% / AUPRO\%).}

    \scriptsize 
    \setlength{\tabcolsep}{3pt}

    \begin{tabular}{cl|cccccccccc|c}
        \hline
        & Method & Bagel & Cable Gland & Carrot & Cookie & Dowel & Foam & Peach & Potato & Rope & Tire & Mean \\
        \hline
        \multirow{9}{*}{\rotatebox{90}{3D}} 
        & VoxelAE \cite{bergmann2021mvtec} & 69.3 / 26.0 & 42.5 / 34.1 & 51.5 / 58.1 & 79.0 / 35.1 & 49.4 / 50.2 & 55.8 / 23.4 & 53.7 / 35.1 & 48.4 / 65.8 & 63.9 / 1.5 & 58.3 / 18.5 & 57.1 / 34.8 \\
        & DepthGAN \cite{bergmann2021mvtec} & 53.0 / 11.1 & 37.6 / 7.2 & 60.7 / 21.2 & 60.3 / 17.4 & 49.7 / 16.0 & 48.4 / 12.8 & 59.5 / 0.3 & 48.9 / 4.2 & 53.6 / 44.6 & 52.1 / 7.5 & 52.3 / 14.3 \\
        & DepthAE \cite{bergmann2021mvtec} & 46.8 / 14.7 & 73.1 / 6.9 & 49.7 / 29.3 & 67.3 / 21.7 & 53.4 / 20.7 & 41.7 / 18.1 & 48.5 / 16.4 & 54.9 / 6.6 & 56.4 / 54.5 & 54.6 / 14.2 & 54.6 / 20.3 \\
        & FPFH \cite{horwitz2022empirical} & 82.5 / \underline{97.3} & 55.1 / 87.9 & 95.2 / 98.2 & 79.7 / 90.6 & 88.3 / 89.2 & 58.2 / 73.5 & 75.8 / 97.7 & 88.9 / 98.2 & 92.9 / 95.6 & 65.3 / 96.1 & 78.2 / 92.4 \\
        & M3DM \cite{wang2023multimodal} & 94.1 / 94.3 & 65.1 / 81.8 & 96.5 / 97.7 & \underline{96.9} / 88.2 & 90.5 / 88.1 & 76.0 / 74.3 & 88.0 / 95.8 & 97.4 / 97.4 & 92.6 / 95.0 & 76.5 / 92.9 & 87.4 / 90.6 \\
        & Shape-Guided \cite{chu2023shape} & \textbf{98.3} / \textbf{97.4} & 68.2 / 87.1 & \underline{97.8} / 98.1 & \textbf{99.8} / \underline{92.4} & \textbf{96.0} / 89.8 & 73.7 / \underline{77.3} & \underline{99.3} / \underline{97.8} & \textbf{97.9} / \textbf{98.3} & 96.6 / 95.5 & 87.1 / \underline{96.9} & 91.6 / \underline{93.1} \\
        & 3DSR \cite{zavrtanik2024cheating} & 94.5 / 92.2 & \underline{83.5} / 87.2 & 96.9 / \textbf{98.4} & 85.7 / 85.9 & 95.5 / \textbf{94.0} & 88.0 / 71.4 & 96.3 / 97.0 & 93.4 / 97.8 & \textbf{99.8} / \textbf{97.7} & \underline{88.8} / 85.8 & \underline{92.2} / 90.7 \\
        & CFM \cite{costanzino2024multimodal} & 94.8 / 96.7 & 77.0 / \textbf{92.2} & 96.8 / 98.1 & 98.1 / \textbf{92.6} & \underline{93.7} / \underline{91.9} & \underline{89.3} / \textbf{96.5} & 69.4 / 96.5 & 90.9 / \underline{98.1} & 93.9 / \underline{96.3} & 81.2 / \textbf{97.6} & 88.5 / \textbf{95.6} \\
        & \textbf{GPAD (Ours)} & \underline{98.2} / 96.4 & \textbf{89.2} / \underline{88.5} & \textbf{100} / \underline{98.3} & 96.2 / 92.2 & 84.0 / 88.5 & \textbf{92.8} / \underline{77.3} & \textbf{99.8} / \textbf{98.1} & \underline{97.5} / \underline{98.1} & \underline{99.5} / 96.1 & \textbf{95.2} / 88.6 & \textbf{95.2} / 92.0 \\
        \hline

        \multirow{6}{*}{\rotatebox{90}{RGB}} 

        & M3DM \cite{wang2023multimodal} & \underline{94.4} / 95.2 & 91.8 / \underline{97.2} & 89.6 / 97.3 & 74.9 / 89.1 & 95.9 / 93.2 & 76.7 / 84.3 & 91.9 / \underline{97.0} & 64.8 / \underline{95.6} & 93.8 / \underline{96.8} & 76.7 / 96.6 & 85.0 / 94.2 \\
        & Shape-Guided \cite{chu2023shape} & 91.1 / 94.6 & \underline{93.6} / \underline{97.2} & 88.3 / 96.0 & 66.2 / \underline{91.4} & 97.4 / \underline{95.8} & 77.2 / 77.6 & 78.5 / 93.7 & 64.1 / 94.9 & 88.4 / 95.6 & 70.6 / 95.7 & 81.5 / 93.3 \\
        & 3DSR \cite{zavrtanik2024cheating} & 84.4 / 92.3 & 93.0 / 97.0 & \underline{96.4} / \textbf{97.9} & 79.4 / 85.9 & \textbf{99.8} / \textbf{97.9} & \underline{90.4} / 89.4 & \underline{93.8} / 94.3 & 73.0 / 95.1 & 97.8 / 96.4 & \underline{90.0} / \textbf{98.0} & 89.8 / 94.4 \\
        & CFM \cite{costanzino2024multimodal} & 93.7 / \underline{96.0} & 86.4 / 96.6 & \textbf{98.4} / \textbf{97.9} & \underline{95.1} / 88.4 & \underline{98.4} / 91.1 & 78.9 / \textbf{91.6} & 91.5 / \textbf{98.1} & 73.6 / \textbf{97.4} & 96.8 / 95.8 & 82.5 / \underline{97.1} & 89.5 / \underline{95.0} \\
        & EasyNet \cite{chen2023easynet} & \textbf{98.2} / 75.1  & \textbf{99.2} / 82.5  & 91.7 / 91.6 & \textbf{95.3} / 59.9  & 91.9 / 69.8 & \textbf{92.3} / 69.9  & 84.0 / 91.7 & \underline{78.5} / 82.7 & \textbf{98.6} / 88.7 & 74.2 / 63.6  & \underline{90.4} / 77.6\\
        
        &\textbf{GPAD (Ours)} & 93.1 / \textbf{96.8} & 87.5 / \textbf{97.4} & 95.3 / \underline{97.5} & 88.2 / \textbf{92.8} & 97.5 / 95.4 & 86.2 / \underline{91.1} & \textbf{97.6} / 96.3 & \textbf{86.3} / 93.5 & \underline{98.5} / \textbf{97.1} & \textbf{93.2} / 92.6 & \textbf{92.3} / \textbf{95.1} \\
        \hline

        \multirow{12}{*}{\rotatebox{90}{RGB+3D}} 
        & VoxelAE \cite{bergmann2021mvtec} & 51.0 / 46.7 & 54.0 / 75.0 & 38.4 / 80.8 & 69.3 / 55.0 & 44.6 / 76.5 & 63.2 / 47.3 & 55.0 / 72.1 & 49.4 / 91.8 & 72.1 / 1.9 & 41.3 / 17.0 & 53.8 / 56.4 \\
        & DepthGAN \cite{bergmann2021mvtec} & 53.8 / 42.1 & 37.2 / 42.2 & 58.0 / 77.8 & 60.3 / 69.6 & 43.0 / 49.4 & 53.4 / 25.2 & 64.2 / 28.5 & 60.1 / 36.2 & 44.3 / 40.2 & 57.7 / 63.1 & 53.2 / 47.4 \\
        & DepthAE \cite{bergmann2021mvtec} & 64.8 / 43.2 & 50.2 / 15.8 & 65.0 / 80.8 & 48.8 / 49.1 & 80.5 / 84.1 & 52.2 / 40.6 & 71.2 / 26.2 & 52.9 / 21.6 & 54.0 / 71.6 & 55.2 / 47.8 & 59.5 / 48.1 \\
        & PatchCore+FPFH \cite{horwitz2022empirical} & 91.8 / 97.6 & 74.8 / 96.9 & 96.7 / 97.9 & 88.3 / 97.3 & 93.2 / 93.3 & 58.2 / 88.8 & 89.6 / 97.5 & 91.2 / 98.1 & 92.1 / 95.0 & 88.6 / 97.1 & 86.5 / 95.9 \\
        & M3DM \cite{wang2023multimodal} & \underline{99.4} / 97.0 & 90.9 / 97.1 & 97.2 / 97.9 & 97.6 / 95.0 & 96.0 / 94.1 & 94.2 / 93.2 & 97.3 / 97.7 & 89.9 / 97.1 & 97.2 / 97.1 & 85.0 / 97.5 & 94.5 / 96.4 \\
        & Shape-Guided \cite{chu2023shape} & 98.6 / 98.1 & 89.4 / 97.3 & 98.3 / \underline{98.2} & 99.1 / \underline{97.1} & 97.6 / 96.2 & 85.7 / \textbf{97.8} & \underline{99.0} / 98.1 & 96.5 / \textbf{98.3} & 96.0 / 97.4 & 86.9 / 97.5 & 94.7 / \underline{97.6} \\
        & 3DSR \cite{zavrtanik2024cheating} & 98.1 / 96.4 & 86.7 / 96.6 & \textbf{99.6} / 98.1 & 98.1 / 94.2 & \textbf{100} / \textbf{98.0} & \textbf{99.4} / 97.3 & 98.6 / 98.1 & \underline{97.8} / 97.7 & \textbf{100} / 97.9 & \textbf{99.5} / \underline{97.9} & \underline{97.8} / 97.2 \\
        & CFM \cite{costanzino2024multimodal} & \underline{99.4} / 97.9 & 88.8 / 97.2 & 98.4 / \underline{98.2} & \underline{99.3} / 94.5 & 98.0 / 95.0 & 88.8 / 96.8 & 94.1 / 98.0 & 94.3 / \underline{98.2} & 98.0 / 97.5 & 95.3 / \textbf{98.1} & 95.4 / 97.1 \\
        &EasyNet \cite{chen2023easynet} & 99.1 / 83.9  & \textbf{99.8} / 86.4  & 91.8 / 95.1 & 96.8 / 61.8  & 94.5 / 82.8 & 94.5 / 83.6  & 90.5 / 94.2 & 80.7 / 88.9 & \underline{99.4} / 91.1 & 79.3 / 52.8  & 92.6 / 82.1\\
        &M3DM-NR \cite{wang2024m3dm} & 97.4 / \textbf{99.3}  & 97.1 / 97.1  & 97.8 / 97.7 & 94.5 / 97.6  & 93.8 / 96.0 & 94.7 / 92.2  & 97.8 / 97.3 & 97.1 / 89.9 & 97.2 / 95.5 & \underline{97.4} / 88.2  & 96.5 / 94.5\\

        & 2M3DF \cite{asad20252m3df} & 99.2 / \underline{98.6} & \underline{96.9} / \textbf{98.3} & \underline{98.8} / \textbf{98.7} & 98.5 / \textbf{98.4} & 98.1 / \underline{96.8} & 94.7 / 94.9 & 97.9 / \textbf{98.6} & 94.2 / 97.1 & 97.6 / \textbf{98.8} & 89.8 / 95.4 & 96.6 / 97.5 \\
        
        & \textbf{GPAD (Ours)} & \textbf{99.8} / 97.8 & 96.8 / \underline{97.8} & \underline{98.8} / 98.0 & \textbf{99.9} / 96.9 & \underline{99.8} / \textbf{98.0} & \underline{98.7} / \underline{97.6} & \textbf{99.9} / \underline{98.3} & \textbf{98.1} / 97.7 & \textbf{100} / \underline{98.0} & 97.2 / 97.0 & \textbf{98.9} / \textbf{97.7} \\
        \hline
    \end{tabular}

\end{table*}

Following multimodal feature fusion and image reconstruction, we design an anomaly segmentation module called PCUnet. In addition to the geometric prior $F_p$, the input of PCUnet includes the output of the general decoder $\hat{I}_a$ and the output of the speific decoder $\hat{I}$, which are concatenated along the channel dimension. PCUnet enhances the standard U-Net \cite{ronneberger2015u} through hierarchical geometric fusion, as shown in Fig.\hyperref[PCUnet]{4}. During the encoding stage, PCUnet extracts four multi-scale features $\{b_1, b_2, b_3, b_4\}$. During the decoding stage, the point cloud features $F_p$ are linearly transformed to match the dimensionality of the image features. These transformed features are then averaged to form a global geometric representation. PCUnet enhances feature recovery and anomaly segmentation by incorporating global geometric information from the point cloud at each decoding layer and injecting geometric prior in the three-layer decoding process. The specific process includes three steps: upsampling, skip connection, and prior injection. These can be expressed as the following operations:

\begin{equation}
F_{dec}^{(l)} = U_l(F_{dec}^{(l+1)}),
\end{equation}
\begin{equation}
F_{dec}^{(l)} = DB_l(Cat(F_{dec}^{(l)}, F_{enc}^{(l)})), 
\end{equation}
\begin{equation}
F_{dec}^{(l)} \leftarrow F_{dec}^{(l)} + P_l(F_p), 
\end{equation}
where $U_l$ denotes bilinear upsampling, $DB_l$ contains two $3\times3$ convolutional blocks with instance normalization, and $P_l$ implements projection of geometric prior.

In the final decoding step, PCUnet employs a convolutional layer to map the fused features into the segmentation space, generating the final anomaly segmentation map $M_{anomaly}$. The calculation of the anomaly score is based on the output result of the PCUnet and the pixel-wise maximum is taken as the anomaly score. Compared with traditional Unet-based segmentation methods, PCUnet has better ability to handle geometric shape defects. This is achieved through the gradual upsampling and deep fusion of geometric features after geometric prior injection.

\subsection{Loss Function}
\label{3.5}

To balance the objectives of feature reconstruction \cite{lin2023latent}, image reconstruction \cite{you2022adtr,liu2024deep}, and segmentation \cite{zavrtanik2024cheating,liu2024deep}, we designed a comprehensive loss function that enhances the model’s performance in complex scenarios. This loss function combines feature reconstruction loss, image reconstruction loss, and segmentation loss.

The feature reconstruction loss measures the ability of the SRN \cite{liu2018linear,zavrtanik2022dsr,zavrtanik2024cheating} to recover features after an anomaly mask is introduced. It is calculated as the mean squared error (MSE) between the reconstructed features $\{F_R^{T},F_R^{S}\}$ generated by the SRN \cite{liu2018linear,zavrtanik2022dsr,zavrtanik2024cheating} and the $\{F_T,F_S\}$. This loss ensures that the model effectively captures critical information from multimodal data at the feature level.

The image reconstruction loss \cite{you2022adtr,liu2024deep} assesses the similarity between the image $\hat{I}$ reconstructed by the speific decoder and the input image $I$ at the pixel level. This ensures that detailed visual information is preserved during fusion and reconstruction while minimizing the impact of noise.

To improve anomaly segmentation accuracy, a focal loss \cite{zavrtanik2024cheating,liu2024deep} is incorporated between the predicted anomaly segmentation mask $M_{anomaly}$ and the ground truth mask $M_{gt}$. The focal loss focuses on difficult-to-segment regions, thereby enhancing segmentation quality, especially at boundaries.

The overall loss function is defined in Equation (13):
\begin{equation}
\begin{split}
    \mathcal{L} = & \, \alpha \cdot \mathcal{L}_{f} \left( F_T, F_S, F_R^{T}, F_R^{S} \right) \\
    & + \beta \cdot \mathcal{L}_{i}\left( I, \hat{I} \right) + \gamma \cdot \mathcal{L}_{s}\left( M_{gt}, M_{anomaly} \right),
\end{split}
\end{equation}
where $\mathcal{L}_{f}$, $\mathcal{L}_{i}$, and $\mathcal{L}_{s}$ represent the feature reconstruction loss, image reconstruction loss, and segmentation loss, respectively. The parameters $\alpha$, $\beta$, and $\gamma$ are set to 1, 10, and 1, respectively, based on empirical findings from related works \cite{zavrtanik2024cheating,zavrtanik2022dsr}.

By optimizing the total loss $\mathcal{L}$, the GPAD model achieves balanced improvements in feature recovery, image reconstruction, and anomaly segmentation, significantly enhancing its performance in complex industrial scenarios.

\section{Experiments}

\subsection{Dataset} 
To evaluate the performance of our method in multimodal industrial anomaly detection, we utilize two widely adopted datasets: MVTec-3D AD \cite{bergmann2021mvtec} and Eyecandies \cite{bonfiglioli2022eyecandies}. The MVTec-3D AD dataset \cite{bergmann2021mvtec} contains 10 categories of industrial objects with 4,147 high-resolution 3D scans and corresponding RGB images. It includes common industrial defects, such as scratches, dents, and holes, and provides pixel-level anomaly annotations. The MVTec-3D AD \cite{bergmann2021mvtec} dataset is widely used in industrial anomaly detection due to its diverse set of object categories and common industrial defects. The Eyecandies dataset \cite{bonfiglioli2022eyecandies} is a synthetic dataset featuring 10 types of candies and sweets, each with varying shapes, materials, and colors, simulating an industrial production setting. It includes RGB images, depth maps, and normal maps with precise annotations for various geometric and texture anomalies. For both datasets, the depth images can be converted into point clouds.

\subsection{Data Processing}
For multimodal feature learning, each sample’s point cloud data undergoes FPS \cite{qi2017pointnet++,wang2023multimodal,chu2023shape} and is divided into $\text{M}$ groups, with each group containing $\text{N}$ points. This fine-grained partitioning captures both the global structure and local geometric details, which enhances the model’s ability to detect anomalies at various scales. We store the point clouds and normal vectors in npz format to facilitate efficient training, enabling quick retrieval and loading. For the Eyecandies dataset \cite{bonfiglioli2022eyecandies}, we use the Uniform Lighting condition, which ensures that the objects are illuminated evenly, minimizing the impact of shadows or strong highlights.

\subsection{Evaluation Metrics} We use three metrics to evaluate our model’s performance in anomaly detection and segmentation. Image-level AUROC (I-AUROC) measures the model’s overall detection accuracy at the image level, while Pixel-level AUROC (P-AUROC) assesses detection accuracy at the pixel level. The Area Under the Per-Region Overlap (AUPRO) quantifies the model’s capability in localizing anomalies at the region level. Together, these metrics provide a comprehensive evaluation of GPAD’s performance across various anomaly scales.

\subsection{Experimental Details}
We implement GPAD using PyTorch 1.13.1. All experiments are conducted on a single NVIDIA RTX 4090, with a batch size of 8 and a total of 200 epochs. Initially, we pre-train the point cloud expert model using point cloud data from all categories in the dataset, randomly selecting 20 samples per category for this step. Subsequently, the pre-trained point cloud expert model is used for geometric prior extraction to train the GPAD model. For each category, we select the best-performing parameters during training.

\begin{table*}
\label{table2}
    \centering
    \caption{Comparison of GPAD and Other Models on the MVTec3D Dataset (P-AUROC\%).}

    \begin{tabular}{ll|cccccccccc|c}
        \hline
        & Method & Bagel & Cable Gland & Carrot & Cookie & Dowel & Foam & Peach & Potato & Rope & Tire & Mean \\
        \hline


        
        & PatchCore+FPFH \cite{horwitz2022empirical} & \textbf{99.6} & 99.2 & 99.7 & \textbf{99.4} & 98.1 & 97.4 & 99.6 & \textbf{99.8} & 99.4 & 99.5 & 99.2 \\
        & M3DM \cite{wang2023multimodal} & 99.5 & 99.3 &99.7 & 98.5 & 98.5 & 98.4 & 99.6 & 99.4 & \textbf{99.7} & \textbf{99.6} & 99.2 \\
        & Shape-Guided \cite{chu2023shape} & - & - & - & - & - & - & - & - & - & - & \textbf{99.6} \\
        & 3DSR \cite{zavrtanik2024cheating} & - & - & - & - & - & - & - & - & - & - & 99.5 \\
        & \textbf{GPAD (Ours)} & \textbf{99.6} & \textbf{99.6} & \textbf{99.9} & 98.7 & \textbf{99.8} & \textbf{99.9} & \textbf{99.8} & 99.4 & 99.5 & \textbf{99.6} & \textbf{99.6} \\
        \hline
    \end{tabular}

\end{table*}

\subsection{Comparative Experiments}

In this section, we present a detailed comparison between GPAD and existing mainstream multimodal anomaly detection methods, evaluating their performance across various anomaly types and complex industrial scenarios. The baseline methods include VoxelAE \cite{bergmann2021mvtec}, DepthGAN \cite{bergmann2021mvtec}, DepthAE \cite{bergmann2021mvtec}, FPFH \cite{horwitz2022empirical}, M3DM \cite{wang2023multimodal}, Shape-Guided \cite{chu2023shape}, 3DSR \cite{zavrtanik2024cheating}, CFM \cite{costanzino2024multimodal}, EasyNet \cite{chen2023easynet}, M3DM-NR \cite{wang2024m3dm} and 2M3DF \cite{asad20252m3df}.

As shown in Table \hyperref[table1]{1}, we conduct experiments with three configurations (3D-based, RGB-based, and 3D+RGB-based) to evaluate GPAD's advantages across different modality combinations, where comparative model data are obtained from their corresponding original publications.

In the 3D-based configuration, while the compared models effectively capture macro-level shape characteristics, they often lack sensitivity to finer geometric details, which can result in missed subtle anomalies. In contrast, GPAD incorporates normal vector features to enhance sensitivity to minor variations, achieving superior detection accuracy through fine-grained, group-level feature extraction.

In the RGB-based configuration, due to the lack of 3D information perception, all methods show low anomaly detection and localization capabilities. Especially for geometric defects such as holes and cuts, the detection ability of RGB-based methods is limited. GPAD has achieved good results for texture-type defects in images with strong robustness, but there is still a large gap in detection ability compared with multi-modal-based methods. 

In the 3D+RGB-based configuration, existing methods \cite{asad20252m3df,bergmann2021mvtec,horwitz2022empirical,wang2023multimodal,chu2023shape,zavrtanik2024cheating,costanzino2024multimodal} employ various feature fusion strategies to enhance multimodal representation. However, these approaches often struggle with suboptimal feature alignment and integration, leading to significant information loss and consequently reduced anomaly detection accuracy. GPAD addresses these issues through a geometry-conditioned attention mechanism, effectively integrating visual and structural information from RGB-D images and point clouds, thus minimizing information loss. This approach enables GPAD to precisely localize anomaly regions and enhances its robustness in handling complex surface textures. 

\begin{figure*}
    \centering
    \label{figure5}
    \includegraphics[width=1\linewidth]{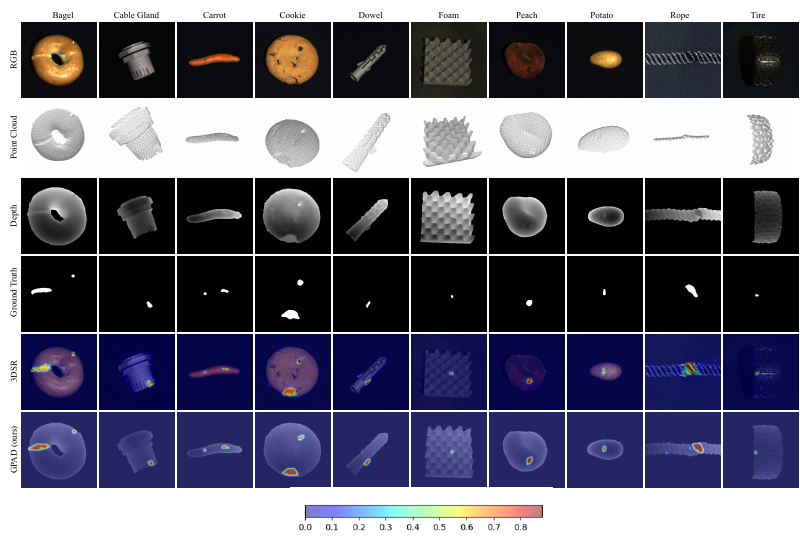}
    \caption{Qualitative results for each class of the MVTec-3D AD \cite{bergmann2021mvtec} dataset. The color bar represents anomaly scores from 0 (normal, blue) to 1 (anomalous, red) using the jet colormap. Compared with 3DSR \cite{zavrtanik2024cheating}, GPAD can obtain a more accurate segmentation region.}
\end{figure*}

As shown in Fig.\hyperref[figure5]{5}, we visualize the heatmap results of our method (GPAD) and 3DSR \cite{zavrtanik2024cheating}, both using multimodal inputs. Compared to the results of 3DSR \cite{zavrtanik2024cheating}, our method demonstrates superior segmentation maps with enhanced multimodal feature representation. Specifically, GPAD provides more accurate anomaly localization, particularly for complex surfaces such as Cable Gland and Tire. It reduces false positives and avoids the over-segmentation seen in 3DSR \cite{zavrtanik2024cheating}. This is attributed to GPAD's geometry-conditioned attention mechanism and the use of geometric prior, which enhance the structural context for better detection. These improvements make GPAD effective for detecting both subtle and complex anomalies.

The results in Table \hyperref[table1]{1} demonstrate GPAD’s superior performance in both 3D-based and 3D+RGB-based configurations, particularly in detecting complex geometric structures and subtle anomalies. In Table \hyperref[table2]{2}, we provide a detailed comparison of the P-AUROC results for the competing models. GPAD achieves an average P-AUROC of 99.6\%, matching the performance of the current best method, Shape-Guided \cite{chu2023shape}. However, the specific P-AUROC values for 3DSR \cite{zavrtanik2024cheating} and Shape-Guided \cite{chu2023shape} are unavailable. We obtain related results on the MVTec-3D AD dataset from the corresponding papers and the \textquotedblleft Papers with Code \textquotedblright platform. To further validate GPAD’s capabilities, additional experiments are conducted on the Eyecandies dataset \cite{bonfiglioli2022eyecandies}, focusing on RGB+3D-based models. The results are presented in Table \hyperref[table3]{3}. GPAD outperforms other models across most categories, achieving a 0.3\% improvement in the I-AUROC metric over the current best method. 

The proposed method, with its innovative structure and integration of geometric prior, has significantly enhanced anomaly detection capabilities, achieving superior performance to state-of-the-art methods on both the MVTec-3D AD \cite{bergmann2021mvtec} and Eyecandies datasets \cite{bonfiglioli2022eyecandies}. However, in the specific task of anomaly regions localization, its performance is comparable to the Shape-Guided \cite{chu2023shape} method, without marked improvement. This situation may arise from two factors. Firstly, the anomaly localization task is highly challenging, demanding precise identification of anomaly locations amidst complex backgrounds and diverse object appearances. Secondly, the current approach to point cloud grouping might not fully leverage the geometric information necessary for accurate localization. Future enhancements could involve further optimizing the point cloud expert model and refining the extraction of geometric prior. By enhancing the model's sensitivity to local geometric features and improving the precision of geometric prior extraction, it is anticipated that the anomaly localization capability of GPAD can be significantly improved.

\begin{table}[t]
    \centering
    \label{table3}
    \caption{Comparison of GPAD and Other Models on the Eyecandies Dataset (I-AUROC\%).}

    \footnotesize
    \setlength{\tabcolsep}{2pt}
    
    \begin{tabular}{l|ccccc}
        \hline
        Category & M3DM \cite{wang2023multimodal} & 3DSR \cite{zavrtanik2024cheating} & CFM \cite{costanzino2024multimodal} & 2M3DF \cite{asad20252m3df} & GPAD (ours) \\
        \hline
        Can.C. & 62.4 & 65.1 & 68.0 & \textbf{75.3} &\underline{71.0} \\
        Cho.C. & 95.8 & \underline{99.8} & 93.1 & 95.5 &\textbf{100} \\
        Cho.P. & \textbf{95.8} & 90.4 & \underline{95.2} & 93.7 &92.8 \\
        Conf. & \textbf{100} & \underline{97.8} & 88.0 & 96.7 &96.0 \\
        Gum.B. & \underline{88.6} & 87.5 & 86.5 & \textbf{90.1} &79.5 \\
        Haz.T. & 75.8 & \textbf{86.1} & 78.2 & 79.2 &\underline{84.8} \\
        Lic.S. & 94.9 & \underline{96.5} & 91.7 & 88.9 &\textbf{98.6} \\
        Lollip. & 83.6 & 89.9 & 84.0 & \textbf{91.3} & \underline{90.8} \\
        Marsh. & \textbf{100} & 99.0 & \underline{99.8} & 98.0 &\textbf{100} \\
        Pep.C. & \textbf{100} & 97.1 & 96.2 & 89.3 &\underline{98.4} \\
        \hline

        Mean & 89.7 & \underline{90.9} & 88.1 & 89.7 &\textbf{91.2} \\
        \hline
    \end{tabular}

\end{table}

\subsection{Ablation Experiments}
\begin{table*}[h]
    \centering
    \label{table4}
    \caption{Performance Comparison of Model Configurations.}

    \begin{tabular}{l|cccccccccc|c}
        \hline
        Config. & Bagel & Cable Gland & Carrot & Cookie & Dowel & Foam & Peach & Potato & Rope & Tire & Mean \\
        \hline

        \multicolumn{12}{c}{I-AUROC} \\ 
        \hline
        w/o Norm       & 99.9 & 95.3 & 100  & 99.5 & 99.5 & 99.4 & 98.6 & 98.0 & 100  & 95.2 & 98.5 \\
        w/o GCA Fusion & 99.7 & 95.0 & 99.9 & 98.3 & 99.6 & 99.0 & 99.5 & 98.3 & 100  & 96.7 & 98.6 \\
        w/o Geo Prior & 99.7 & 95.0 & 99.7 & 99.5 & 99.7 & 99.6 & 98.8 & 98.7 & 99.7 & 97.1 & 98.7 \\
        w/o Depth & 99.3 & 93.7 & 99.4 & 98.7 & 99.2 & 98.7 & 99.2 & 98.3 & 99.1 & 97.5 & 98.3 \\
        w/o Points & 99.1 & 92.7 & 99.3 & 98.8 & 99.5 & 98.4 & 97.4 & 97.1.2 & 99.7 & 96.1 & 97.8 \\

        GPAD & 99.8 & 96.8 & 98.8 & 99.9 & 99.8 & 98.7 & 99.9 & 98.1 & 100 &  97.2 & \textbf{98.9}\\
        \hline        
        \multicolumn{12}{c}{P-AUROC} \\ 
        \hline
        w/o Norm       & 99.6 & 98.8 & 99.9 & 98.6 & 99.6 & 99.8 & 99.8 & 99.6 & 100  & 99.6 & 99.5 \\
        w/o GCA Fusion & 99.6 & 99.3 & 99.9 & 96.8 & 99.6 & 99.6 & 99.8 & 99.7 & 100  & 99.7 & 99.4 \\
        w/o Geo Prior & 99.7 & 98.0 & 99.9 & 99.0 & 99.7 & 99.7 & 99.6 & 99.7 & 99.9 & 99.6 & 99.4 \\
        w/o Depth & 99.9 & 98.5 & 99.9 & 98.6 & 99.5 & 99.7 & 99.6 & 99.2 & 98.9 & 99.4 & 99.3 \\
        w/o Points & 99.6 & 99.4 & 99.9 & 98.8 & 99.7 & 99.9 & 99.1 & 99.6 & 99.0 & 99.6 & 99.5 \\

        GPAD & 99.6 & 99.6 & 99.9 & 98.7 & 99.8 & 99.9 & 99.8 & 99.4 & 99.5 & 99.6 & \textbf{99.6} \\
        \hline
        \multicolumn{12}{c}{AUPRO} \\ 
        \hline
        w/o Norm       & 97.5 & 96.8 & 98.2 & 96.1 & 98.0 & 97.4 & 98.4 & 97.9 & 97.9 & 97.7 & 97.6 \\
        w/o GCA Fusion & 97.4 & 94.0 & 98.1 & 91.3 & 97.7 & 94.8 & 98.3 & 98.1 & 98.0 & 97.2 & 96.5 \\
        w/o Geo Prior & 97.6 & 95.8 & 98.1 & 96.6 & 98.0 & 97.5 & 98.1 & 98.2 & 98.1 & 97.7 & 97.6 \\
        w/o Depth & 97.1 & 97.5 & 98.1 & 95.4 & 97.8 & 97.3 & 97.9 & 97.4 & 97.5 & 96.5 & 97.3 \\
        w/o Points & 97.9 & 97.4 & 98.2 & 96.3 & 98.1 & 97.6 & 97.6 & 97.9 & 97.9 & 97.5 & 97.2 \\

        GPAD & 97.8 & 97.8 & 98.0 & 96.9 & 98.0 & 97.6 & 98.3 & 97.7 & 98.0 & 97.0 & \textbf{97.7} \\
        \hline
        
    \end{tabular}

\end{table*}
\begin{table*}[h]

    \centering
    \label{table5}
    \caption{Comparison of Results for Different Group Quantities in Point Cloud Samples.}

    \begin{tabular}{l|cccccccccc|c}
    \hline
    Num & Bagel & Cable Gland & Carrot & Cookie & Dowel & Foam & Peach & Potato & Rope & Tire & Mean \\
    \hline

    \multicolumn{12}{c}{I-AUROC} \\
    \hline
    100 & 99.7 & 92.8 & 99.9 & 99.3 & 99.6 & 99.4 & 99.3 & 98.2 & 100  & 99.4 & 98.8 \\
    200 & 99.8 & 96.8 & 98.8 & 99.9 & 99.8 & 98.7 & 99.9 & 98.1 & 100  & 97.2 & \textbf{98.9} \\
    300 & 99.7 & 96.7 & 99.4 & 99.3 & 99.6 & 99.5 & 99.5 & 98.5 & 100  & 97.2 & \textbf{98.9} \\
    400 & 99.8 & 96.1 & 100  & 98.9 & 99.9 & 99.2 & 99.6 & 98.4 & 99.7 & 95.9 & 98.8 \\
    500 & 99.8 & 93.7 & 99.8 & 99.1 & 99.6 & 99.1 & 99.5 & 98.8 & 100  & 96.5 & 98.6 \\
    \hline
    \multicolumn{12}{c}{P-AUROC} \\
    \hline
    100 & 99.7 & 97.3 & 99.9 & 98.7 & 99.7 & 99.8 & 99.6 & 99.6 & 99.4 & 99.8 & 99.4 \\
    200 & 99.6 & 99.6 & 99.9 & 98.9 & 99.6 & 99.9 & 99.8 & 99.4 & 99.5 & 99.6 & \textbf{99.6} \\
    300 & 99.7 & 99.4 & 99.9 & 98.4 & 99.5 & 99.8 & 99.2 & 99.7 & 99.5 & 99.6 & 99.5 \\
    400 & 99.6 & 98.6 & 100  & 96.7 & 99.7 & 99.7 & 99.7 & 99.6 & 99.2 & 99.4 & 99.2 \\
    500 & 99.5 & 99.1 & 99.9 & 98.9 & 99.7 & 99.8 & 99.8 & 99.8 & 99.2 & 99.5 & 99.5 \\
    \hline
    \multicolumn{12}{c}{AUPRO} \\

    \hline
    100 & 97.7 & 94.6 & 98.3 & 96.4 & 98.1 & 97.6 & 98.2 & 97.7 & 97.9 & 97.7 & 97.4 \\
    200 & 97.8 & 97.8 & 98.0 & 96.9 & 98.0 & 97.6 & 98.3 & 97.7 & 98.0 & 97.0 & \textbf{97.7} \\
    300 & 97.2 & 97.0 & 98.1 & 95.8 & 97.9 & 97.4 & 97.8 & 98.1 & 98.0 & 97.0 & 97.4 \\
    400 & 97.9 & 95.0 & 98.3 & 92.9 & 97.9 & 96.9 & 98.3 & 97.9 & 97.9 & 97.2 & 97.0 \\
    500 & 97.3 & 96.3 & 98.2 & 96.1 & 98.1 & 97.7 & 98.4 & 98.1 & 97.8 & 97.2 & 97.5 \\
    \hline
    
    \end{tabular}

\end{table*}

To assess the contribution of each module and design strategy in GPAD, we conduct ablation experiments to evaluate the impact of normal vectors, geometry-conditioned attention fusion, and geometric prior on model performance. These experiments are performed on the MVTec-3D AD \cite{bergmann2021mvtec} dataset using I-AUROC, P-AUROC, and AUPRO as metrics to measure anomaly detection and segmentation capabilities under different configurations. The model configurations are as follows: (1) w/o Norm: removing the normals encoding from the point cloud expert model, using only original point cloud data without normal vectors. (2) w/o GCA Fusion: removing the “geometry-conditioned attention fusion”, using geometric prior in PCUnet. (3) w/o Geo Prior: excluding geometric prior from the anomaly segmentation module, using \textquotedblleft geometry-conditioned attention fusion \textquotedblright. (4) w/o Depth: remove the depth image. (5) w/o Points: using only RGB and depth images. This ablation framework isolates the impact of each component for rigorous effectiveness validation.

As shown in Table \hyperref[table4]{4}, quantitative results demonstrate clear performance degradation when removing any key component. Removing normal vectors (w/o Norm) leads to a noticeable drop in detection accuracy, particularly for subtle anomalies, as normal vectors capture essential geometric details of the object surface. The absence of geometry-conditioned attention fusion (w/o GCA Fusion) results in suboptimal feature alignment and integration, reducing the model's ability to fully exploit the complementary nature of RGB and 3D data and the ability to localize anomalies. Similarly, excluding geometric prior (w/o Geo Prior) diminishes the model's ability to leverage 3D structural context, leading to lower segmentation accuracy. Removing the depth image (w/o Depth) affects the early geometric perception of the RGB image, resulting in a drop in detection performance for many categories. When removing the points encoding (w/o Points), due to the characteristics of image data, this will reduce the model's spatial perception of the object structure, resulting in a significant decrease in the detection results. These findings underscore the critical role of each module in GPAD's architecture. The depth image provides early geometric perception for the RGB image, the normal vectors enable fine-grained detection of surface variations, the geometry-conditioned attention fusion ensures effective integration of multimodal features, and the geometric prior provides valuable structural context for precise anomaly segmentation. Through the single-modal and multimodal detection results in Table \hyperref[table1]{1}, and the results of the ablation experiment in Table \hyperref[table4]{4}, the detection results of three modalities are significantly better than those of fewer modalities, and these components together contribute to GPAD's leading performance in industrial anomaly detection.


\subsection{Analysis of Group Quantity}
\begin{figure}
    \centering
    \includegraphics[width=1\linewidth]{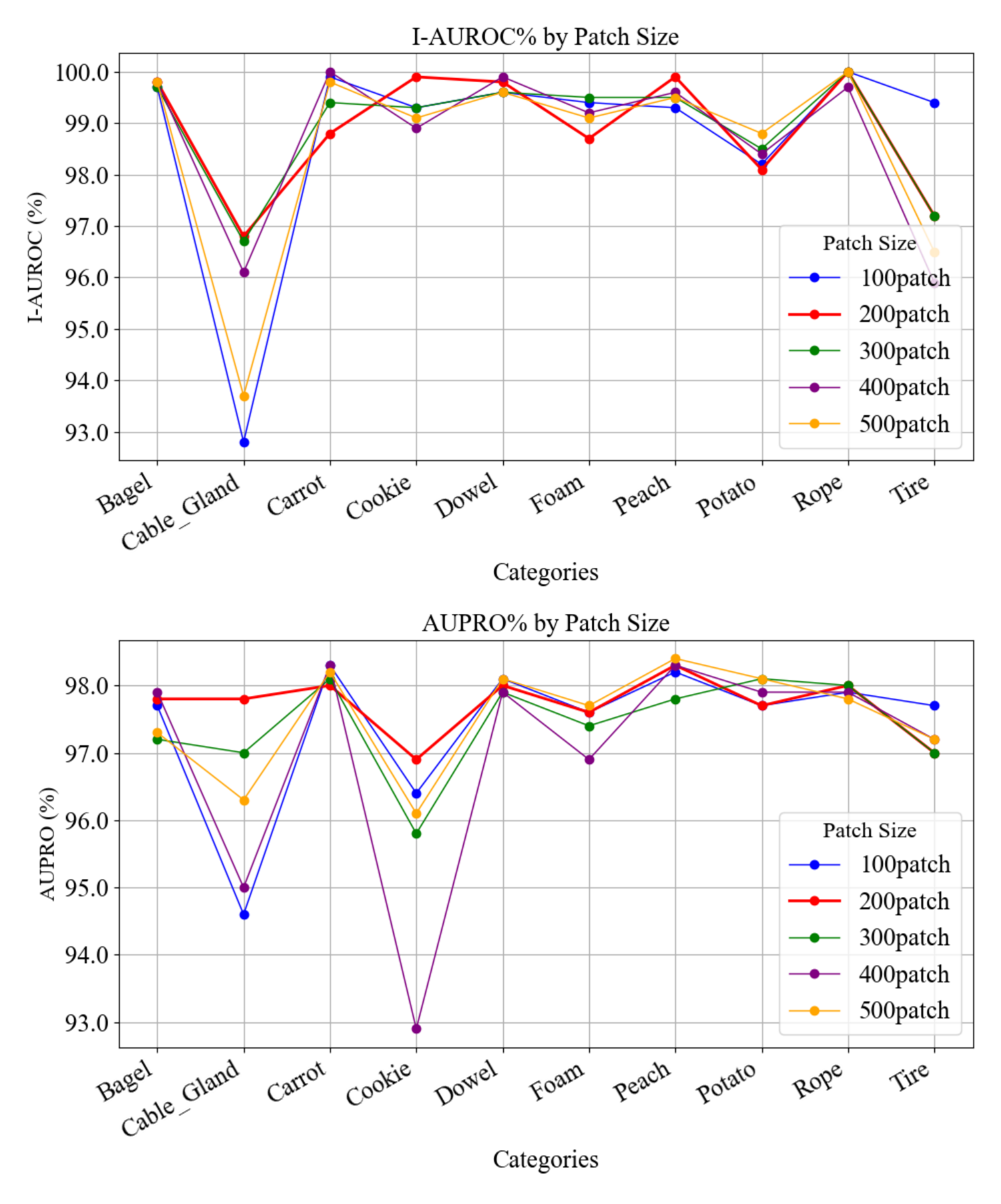}
    \caption{Comparison of detection results (I-AUROC and AUPRO) on the MVTec-3D AD dataset when each sample is divided into 100, 200, 300, 400, and 500 groups.}
    \label{figure6}
\end{figure}
In GPAD's design, the granularity of point cloud partitioning plays a crucial role in model performance. Specifically, each point cloud sample is divided into multiple groups \cite{qi2017pointnet++,wang2023multimodal,chu2023shape}, with each group containing a fixed number of points, allowing for detailed extraction of local geometric features. This grouping method divides the entire point cloud into several localized regions based on spatial distribution, ensuring comprehensive coverage of local geometric structures and effectively enhancing feature representation.

In this experiment, we aim to balance feature extraction precision with computational efficiency by setting the number of points per group to $\text{N}$ = 500. The total number of groups $\text{M}$ directly influences the granularity of the point cloud, thereby affecting the model's ability to capture geometric details. We evaluate GPAD's performance with different numbers of groups $\text{M}$ (set to 100, 200, 300, 400, and 500), as illustrated in Fig.\hyperref[figure6]{6}. 

To clearly demonstrate the impact of group quantity on model performance, we calculate the average results for I-AUROC, P-AUROC, and AUPRO metrics across different configurations, which are summarized in Table \hyperref[table5]{5}. Some categories, such as Bagel, Carrot, Cookie, Dowel, Foam, Potato, Rope, and Peach, exhibit relatively stable performance regardless of the grouping quantity. This is because the surfaces of these objects are relatively smooth, and both larger and smaller quantities of point cloud groups effectively describe surface variations. However, categories like Cable Gland and Tire have more complex surface structure. Excessive point cloud groups introduce unnecessary noise and increase computational overhead, while too few groups fail to capture sufficient geometric information, resulting in reduced accuracy. From Table \hyperref[table5]{5}, we observe a general trend: increasing the number of groups improves GPAD's performance across all metrics, especially when $\text{M}$ is set to 200 or 300, where performance gains are most notable. However, as the number of groups exceeds 300, the improvements plateau, and both I-AUROC and AUPRO show slight declines at 400 and 500 groups. This suggests that while a moderate increase in group quantity enhances fine-grained feature extraction, an excessively high number of groups introduces additional computational complexity and redundancy, which negatively impacts model efficiency. Therefore, selecting 200 groups strikes an optimal balance, ensuring both strong performance and efficient computation.

\subsection{Inference Efficiency and Memory Consumption}
\begin{table}[t]
    \centering
    \label{inference}
    \caption{Inference speed and memory consumption comparison.}

    \scriptsize

    \begin{tabular}{lccccc}
        \hline
        \multirow{2}{*}{Method} & Memory Consumption& Inference & Mean \\
              & (MBs) & (FPS) & I-AUROC(\%) \\

        \hline
        BTF \cite{horwitz2023back} & 228.984 & 3.91 & 86.5  \\
        AST \cite{rudolph2023asymmetric} & 463 & 4.78 & 93.7 \\
        Shape-Guided \cite{chu2023shape} & 237.58 & 2.1 & 94.7  \\
        M3DM \cite{wang2023multimodal} & 6528.7 & 0.514 & 94.5 \\
        CPMF \cite{cao2024complementary} & 2195 & 0.609 & 95.1  \\
        2M3DF \cite{asad20252m3df} & 578.46 & 29.8 & 96.6  \\

       \textbf{GPAD(Ours)} & \textbf{708.5} & \textbf{25.8} & \textbf{98.9}  \\

        \hline
    \end{tabular}

\end{table}

Due to the high requirements for real-time performance and hardware in actual industrial production environments, in this section, we further evaluate the frames-per-second (FPS) and memory consumption of the proposed model. As shown in Table \hyperref[inference]{6}, we compare the FPS and memory consumption of GPAD with those of some recent papers. The experimental platform also uses the NVIDIA RTX 4090 GPU, and the data in the table comes from the 2M3DF \cite{asad20252m3df} paper.

The experimental results show that due to the introduction of the point cloud expert model, the computational complexity of this part is 1.15GFLOPs, and GPAD is slightly higher than the 2M3DF \cite{asad20252m3df} method in terms of memory consumption. In terms of inference speed, although the FPS of 2M3DF \cite{asad20252m3df} is 0.16 times higher than that of GPAD, GPAD improves the false detection rate by 2.1 times by introducing geometric prior, better balancing detection performance and computational efficiency. Compared with previous methods, such as M3DM \cite{wang2023multimodal}, which uses PointTransformer as a point cloud feature extractor, the computational complexity of the point cloud processing part is 9.64 GFLOPs. GPAD far exceeds such memory-based methods in terms of memory consumption, detection speed, and detection accuracy. These findings demonstrate the strong potential of GPAD for practical industrial applications, where real-time performance and high detection accuracy are crucial.

\section{Limitation and Future Work}

The proposed model faces challenges in fully meeting the real-time and hardware requirements of industrial scenarios. In future research, we will further optimize the computational efficiency of the model and explore lightweight designs to achieve real-time detection in production environments, thereby promoting its wider application in industrial detection. Specifically, we have several specific plans: Firstly, we will train a lightweight student network through knowledge distillation to reduce the complexity of the point cloud expert model. To achieve this, we will adopt a teacher-student framework where the teacher network is our current point cloud expert model, and the student network is a simplified version with fewer parameters. We will use the Distilling Dark Knowledge method to transfer knowledge from the teacher network to the student network. Secondly, we will explore model pruning techniques to remove redundant parameters and connections. We will use the magnitude-based pruning method, which iteratively removes weights with the smallest magnitudes and retrained the pruned model to fine-tune the remaining weights. This process will be repeated until we achieve a balance between model size and performance. Through these efforts, we hope to enhance the adaptability of the model to industrial applications.

\section{Conclusion}

In this paper, we present GPAD, a geometric prior-guided multimodal anomaly detection model. We propose a point cloud expert model to extract geometric prior knowledge, addressing the issue of geometric distortion through a fusion mechanism based on geometric prior. Additionally, we utilize geometric prior to guide the segmentation of abnormal regions, enhancing the geometric perception ability of the segmentation network. Experimental results demonstrate that GPAD significantly outperforms current mainstream methods on industrial datasets such as MVTec-3D AD and Eyecandies. Overall, the successful application of GPAD in multimodal anomaly detection highlights its great potential in complex industrial scenarios. 




\vspace{-33pt}

\begin{IEEEbiography}[{\includegraphics[width=1in,height=1.25in,clip,keepaspectratio]{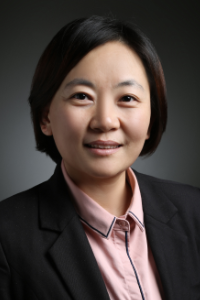}}]{Min~Li}
is a professor of software engineering at Qilu University of technology (ShanDong Academic of Sciences). She received her master’s degree in Communications Engineering from Tianjin University. Her research interests include information technology standardization, economic and information development, big data analysis and application, data governance and data openness, digital government planning and evaluation.
\end{IEEEbiography}
\vspace{-33pt}

\begin{IEEEbiography}[{\includegraphics[width=1in,height=1.25in,clip,keepaspectratio]{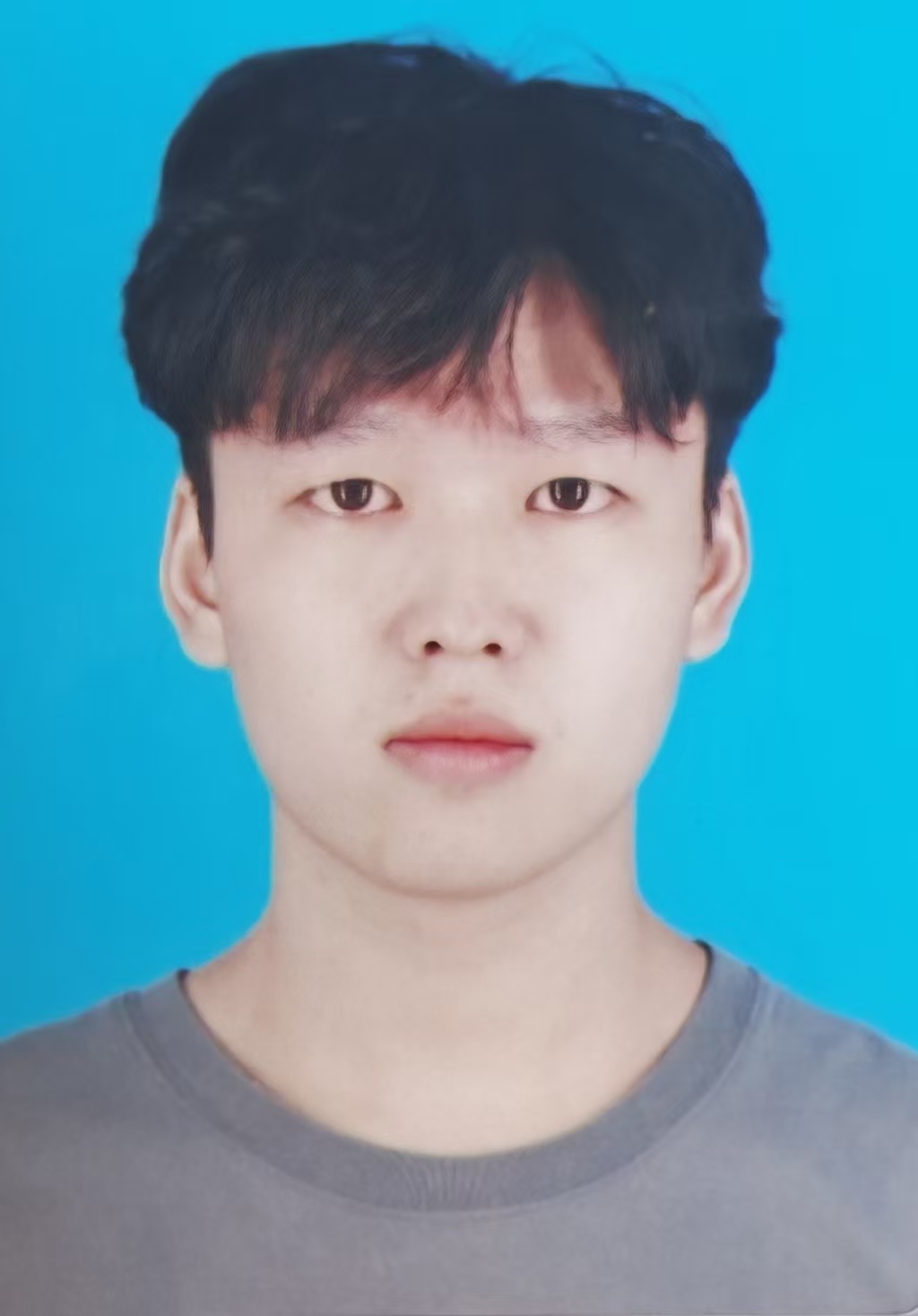}}]{Jinghui~He}
is currently pursuing the M.S. degree with the School of Computer Science and Technology, Qilu University of Technology (Shandong Academy of Sciences), Jinan, China. His current research interests include computer vision, multimodal anomaly detection and 3D anomaly detection.
\end{IEEEbiography}
\vspace{-33pt}

\begin{IEEEbiography}[{\includegraphics[width=1in,height=1.25in,clip,keepaspectratio]{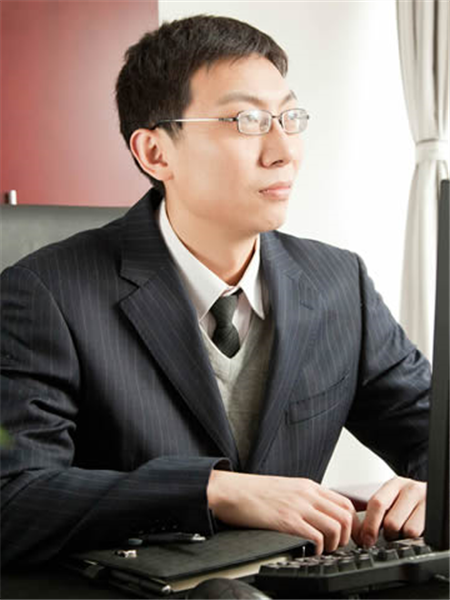}}]{Gang~Li}
received the Ph.D. degree in Management Science and Engineering from Harbin Institute of Technology, Harbin, China. He is currently a Full Professor of the School of Computer Science and Technology, Qilu University of Technology (Shandong Academy of Sciences), and a young expert of Mount Taishan Scholars. His current research interests include machine vision, pattern recognition, large model, big data analysis and application.
\end{IEEEbiography}

\vspace{-33pt}

\begin{IEEEbiography}[{\includegraphics[width=1in,height=1.25in,clip,keepaspectratio]{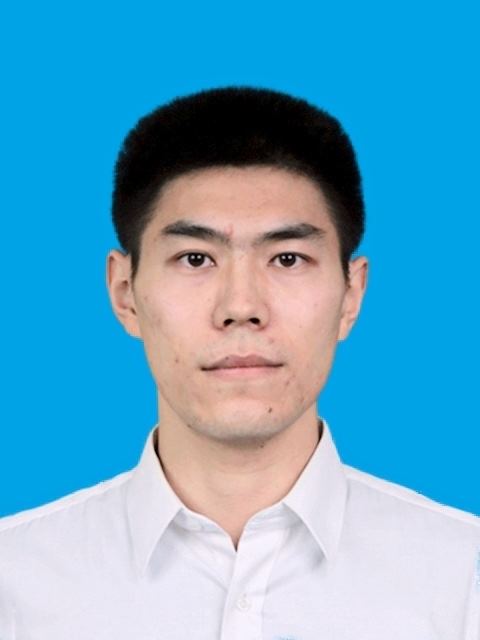}}]{Jiachen~Li}
is an associate professor of the School of Computer Science and Technology, Qilu University of Technology (Shandong Academy of Sciences). He received his Ph.D. degree from Shandong University, Jinan, China. His research interests include augmented reality, 3D object tracking, reconstruction, anomaly detection, etc.
\end{IEEEbiography}
\vspace{-33pt}

\begin{IEEEbiography}[{\includegraphics[width=1in,height=1.25in,clip,keepaspectratio]{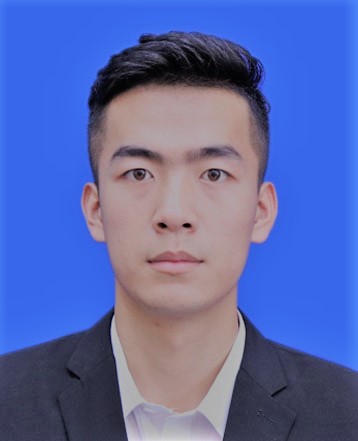}}]{Jin~Wan}
received his Ph.D. degree from Beijing Jiaotong University, Beijing, China, in 2023. He is currently an associate professor with the School of Computer Science and Technology, Qilu University of Technology (Shandong Academy of Sciences). His current research interests include computer vision, pattern recognition, and signal processing.
\end{IEEEbiography}

\vspace{-33pt}
\begin{IEEEbiography}[{\includegraphics[width=1in,height=1.25in,clip,keepaspectratio]{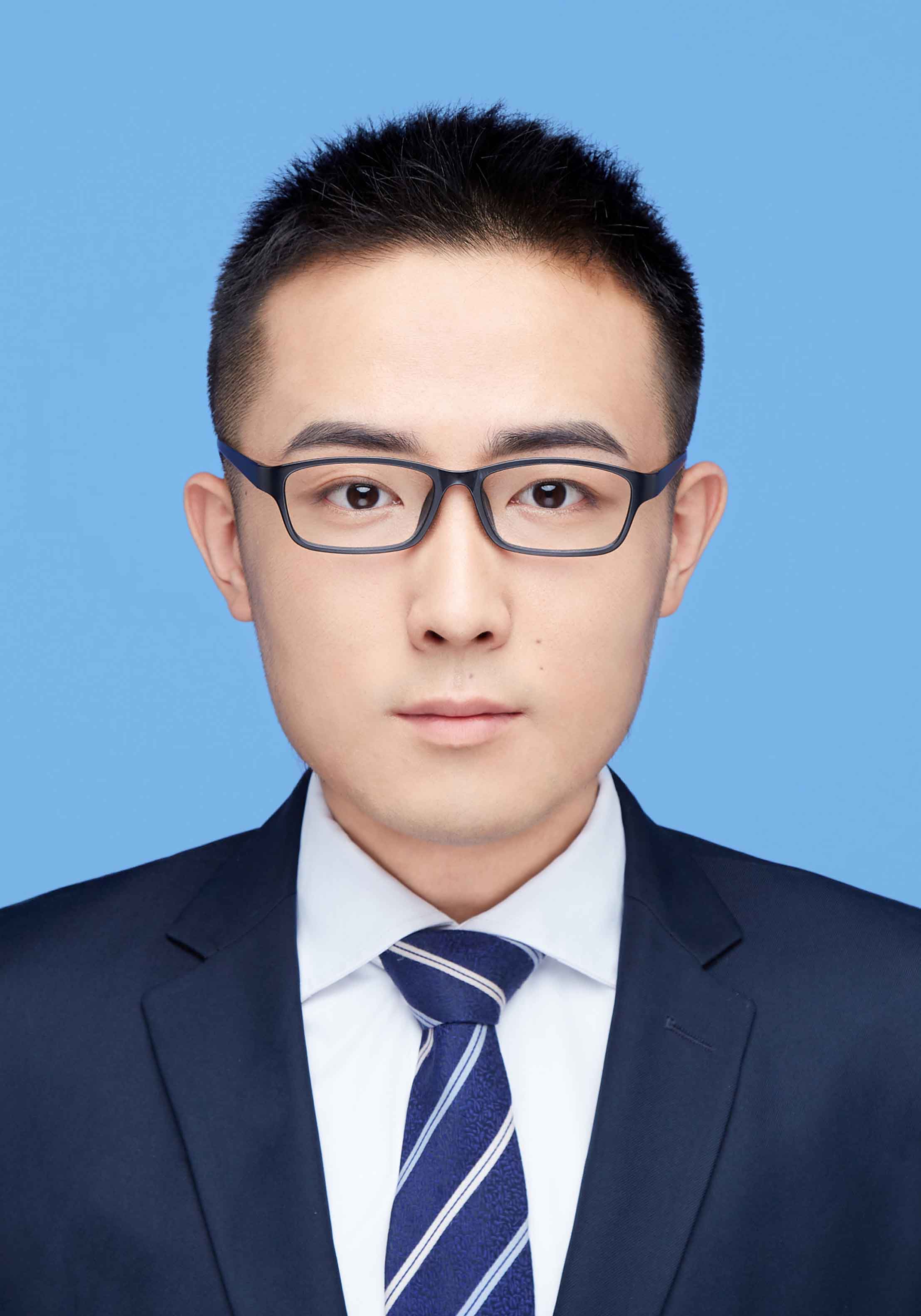}}]{Delong~Han}
received the Ph.D. degree in Electronic Science and Technology from Beijing University of Posts and Telecommunications, Beijing, China. He is currently a research associate of the School of Computer Science and Technology, Qilu University of Technology (Shandong Academy of Sciences). His current research interests include machine vision, pattern recognition, and digital government.
\end{IEEEbiography}

\end{document}